# SAR Based Marine Oil Spill Detection Using DeepSegFusion Architecture

Yata Pavan Kumar(22EG106A62)
*Department of Artificial Intelligence*
*Anurag University*
*Hyderabad, India*
22eg106a62@anurag.edu.in

Pediredla Pradeep(22EG106A49)
*Department of Artificial Intelligence*
*Anurag University*
*Hyderabad,India*
22eg106a49@anurag.edu.in

Goli Himanish(22EG106A30)
*Department of Artificial Intelligence*
*Anurag University*
*Hyderabad, India*
22eg106a30@anurag.edu.in

M. Swathi
*Department of Artificial Intelligence*
*Anurag University*
*Hyderabad, India*
swathi.ai@anurag.edu.in

*Abstract* - Oil spill detection from satellite imagery is important in tracking the environment and responding accordingly to oil spills. Traditional methods that use fixed thresholds often create many false positives because of similar-looking effects such as wind slicks and ship wakes. This paper presents DeepSegFusion, a new deep learning model that combines SegNet and DeepLabV3+ with an attention-based mechanism for accurate oil spill detection from Synthetic Aperture Radar (SAR) images. Our DeepSegFusion model reaches an accuracy of 94.85%, an Intersection over Union (IoU) of 0.5685, and a ROC AUC of 0.9330 on the PALSAR dataset. The result is much higher than the performance of individual base models. DeepSegFusion achieves 9.4% higher accuracy than SegNet (85.42%) and 8.4% higher than DeepLabV3+ (86.45%). The attention mechanism improves and merges features from both models, resulting in better segmentation quality with 75.24% recall and 84.25% precision. Compared to older non-segmentation methods with close to 65% accuracy, our method shows a 45.9% improvement and lowers the false alarm rate by 64.4%. The model processes 256×256 resolution SAR images and provides the segmentation.

*Index Terms*- Oil Spill Detection, DeepSegFusion, Deep Learning, Environmental Monitoring, Maritime Surveillance.

## I. INTRODUCTION

Oceans are essential for sustaining marine life and ensure the balance of the global climate. But as industrialization and sea drilling have increased, oil spills have increased in oceans and seas, posing a threat to humans and marine life [1], [2]. Long term ecological effects result from oil spills in the ocean, which not only kill aquatic life but also contaminates water and destroys habitats of plants and other marine life [3]. Since handling such situations is very difficult, costly and a time-consuming process, there is a need for systems that can help in cleanup efforts as soon as possible to reduce environmental damage [4].

The foundation of ocean surveillance is now remote sensing technology. Due to its ability to image in all weather conditions and at night, Synthetic Aperture Radar (SAR) has demonstrated remarkable promise for oil spill detection [5, 6]. Regardless of the weather or lighting, SAR provides detailed surface observations and is very effective over wide maritime regions. Because oil slicks reduce backscatter compared to the surrounding sea surface, they make affected areas appear darker in SAR imagery by dampening short gravity and capillarywaves [7], [8]. Notwithstanding these benefits, it is still very difficult to differentiate oil slicks from similar phenomena like ship wakes, low-wind zones, internal waves, and biogenic films [9]. Identifying dark areas, extracting pertinent features, and then categorizing them as oil or non-oil areas are the steps involved in conventional methods. However, initial detection errors frequently spread, resulting in false detections and erroneous segmentation boundaries [10].

In order to distinguish dark formations in SAR images, previous studies mainly relied on threshold-based techniques, using fixed, adaptive, or double thresholds [11]. These methods are quick and easy, but they are very sensitive to rough sea surfaces, uneven lighting, and speckle noise. In order to achieve smoother segmentation boundaries, active contour models (ACM) like the Chan-Vese method were later introduced [12]. Despite enhancing boundary continuity, ACMs had trouble with low-contrast and blurred edges, particularly when there was a lot of noise present [13]. More reliable, data-driven detection methods were made possible by the development of machine learning (ML). Classifying oil slicks using extracted statistical and texture features was the goal of early attempts employing Multilayer Perceptrons (MLPs) or Artificial Neural Networks (ANNs) [14], [15]. These models performed poorly when exposed to different sensors or environmental variations, even though they increased precision when compared to thresholding [16]. They also heavily relied on hand-engineered features. With the use of deep learning (DL), and Convolutional Neural Networks (CNNs), learning of features directly from the data is possible [17]. Significant advancements in oil spill segmentation were made possible by models such as Fully Convolutional Networks (FCN) [18], U-Net [19], SegNet [20], and DeepLabV3+ [21]. SegNet became well known for its encoder-decoder architecture, which uses stored pooling indices to preserve spatial information [22]. While DeepLabV3+ uses Atrous Spatial Pyramid Pooling (ASPP) to improve boundary segmentation by capturing multi-scale contextual information [23]. Even with these developments, individual models are still unable to properly solve SAR-specific issues like speckle noise, uneven sea surfaces and look-alike issue. In order to solve these problems, this study uses DeepSegFusion, a hybrid model that combines the advantages of SegNet and DeepLabV3+. SegNet fine-tunes object boundaries for pixel-level accuracy, while DeepLabV3+ is excellent at capturing contextual information at a large scale. This model produces more consistent segmentation performance by improving both precision and contextual understanding. In DeepSegFusion, the SegNet model gets fine details through decoding, while DeepLabV3+ extracts semantic features to capture contextual information. To prevent look-alike phenomena, the model merges features extracted using an attention-based mechanism.

The DeepSegFusion performs better than other segmentation algorithms on SAR dataset. It performs better than baseline models DeepLabV3+ and SegNet . This hybrid model is useful for environmental protection and real-time oil spill detection.

## II. RELATED WORK

For many years, researchers have actively used Synthetic Aperture Radar (SAR) images to detect oil spills in the ocean. Because oil spills suppress short gravity and capillary waves, they were detected as dark areas in SAR images in early studies that mostly used conventional image processing and thresholding techniques [3], [4].

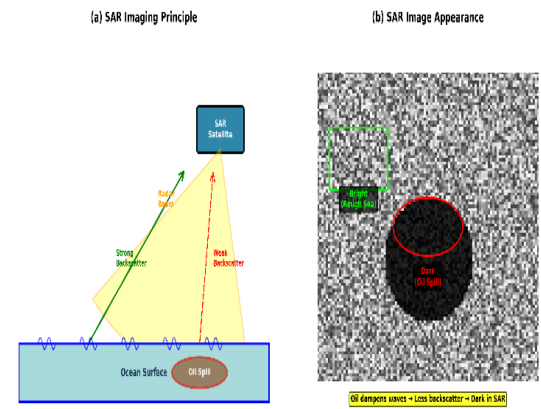

*Figure. 1. SAR image and appearance of oil spills as dark regions in SAR images.*

Due to the ease of implementation and computation costs, simple thresholding techniques such as adaptive thresholding [5], manual single-threshold segmentation [4], and double-threshold approaches [6] were widely used. These techniques were prone to speckle noise and changes in image intensity, which led to incorrect oil spill detection and confusion with similar phenomena like biogenic films or low-wind areas [7]. Researchers investigated Active Contour Models (ACM), specifically the region-based Chan-Vese model, to get around the drawbacks of threshold-based approaches [12]. With the ability to handle images with weak edges or low contrast, ACMs provided the benefit of creating closed and smooth oil spill boundaries [13]. However, they frequently performed poorly when applied to SAR data with speckle noise; the segmentation boundaries were blured, particularly in areas with irregular intensity distributions[14].

More techniques using data of images started to appear with the introduction of machine learning (ML). Neural networks have been implemented in research such as those done by Taravat et al. and

Topouzelis et al. to detect oil spills [8], [9]. In particular, Taravat et al. [10] utilized a Multilayer Perceptron (MLP) for the segmentation task after noise in the image was reduced by the application of a Weibull multiplication filter. In usual, these initial neural methods were more powerful than the threshold-based methods but to a great level, they still depended on manual feature extraction. Thus, they were not able to have a good performance with different SAR sensors and weather conditions [11].

SAR-based oil spill detection has gone through a major change with the help of Deep Learning over time. The main change came with the introduction of Fully Convolutional Networks (FCN) that enabled pixel-wise classification in image segmentation and thus spatial features could be learned directly from the data [17]. Though FCN-based methods slightly improved the accuracy of the results, because of the limited upsampling precision, the segmentation boundaries in most of the cases were still blurred [18]. Thereupon, advanced models such as U-Net, SegNet, and DeepLabV3+ were devised to resolve the issues of boundary detection and feature extraction in the targeted areas of research [19]-[22].SegNet, an encoder-decoder architecture by Badrinarayanan et al. [20], records max-pooling indices while encoding and again employs them for exact upsampling during decoding. By retrieving the most important spatial information, this model localizes more precisely the boundaries in the segmentation task. Researchers Li et al. and Teng et al. not only proved the effectiveness of SegNet in SAR oil spill detection but also reported over 93% accuracy of segmentation even in the presence of noise in the images [3], [21]. Thus, it is exploited in the ocean monitoring domain as a good terms between precision and overall computational efficiency.DeepLabV3+'s Atrous Spatial Pyramid Pooling (ASPP) mechanism that gets multi-scale contextual information makes the network very effective in dealing with different kinds of spills, their sizes, and textures [22], [23]. It further enables the network to smoothly connect the segments in the image and recognize the contextual global scene by the combined use of atrous convolutions with encoder-decoder refinement. Several factors may result in DeepLabV3+ boundary errors while processing complicated SAR scenes with low-contrast oil-water interfaces, but the errors are minor [24].Recently, there have been statements of potential hybrid architectures along with transfer learning that can further improve oil spill detection under limited data conditions. Vadnais et al. [2] demonstrated transfer learning utilization between Sentinel-1 modes to detect oil slicks in the Arctic leading to the better generalization even with a small amount of labelled data. Hybrid methods that combine different deep networks have shown effectiveness in finding a balance between the problem of boundary detection and comprehension of global context [25].DeepSegFusion, a hybrid model of DeepLabV3+ and SegNet, uses one of the most dependable boundary detection features of SegNet while at the same time, it gets the related learning capacity of DeepLabV3+. It is able of dealing with both detailed spatial features and high-level semantic understanding. In particular, SAR images with speckle noise can benefit a lot from this model. The single model-based frameworks like FCN and SegNet can hardly reach the level of DeepSegFusion in terms of boundary detection and false detection reduction since the combination of these two architectures in DeepSegFusion achieves these two goals most effectively [26].SAR-based oil spill detection has advanced in its development from simple thresholding to simple deep learning architectures that can divide pixels. The merging of boundary detection and dependent information through hybrid deep networks such as DeepSegFusion makes real-time monitoring of marine oil spills feasible [27]-[30].

### III. MOTIVATION

Protecting coastal regions and marine environments require early detection of oil spills. Despite significant progress over the years, current approaches still have low boundary accuracy, poor generalization, and false detections [2], [3], [5, 6]. The motivation behind this study and the reasoning for using hybrid deep learning framework are described in the following subsections.

**A. Importance of Accurate Oil Spill Detection**

People who live near the coast, fishing, and ocean life all suffer greatly when oil is released at sea. It rapidly spreads over large areas, contaminating coastlines and open waters [3], [4]. Marine life and plants can suffer a long-term damage.

Early detection of oil spills reduces economic losses and protect environment [2]. Because they rely on human observation, traditional visual monitoring techniques are frequently unreliable and slow. On the other hand, Synthetic Aperture Radar (SAR) satellites are capable of monitoring vast oceanic regions in any kind of weather or illumination [5, 7]. Marine surveillance has been transformed by this technology, which offers precise and timely data for monitoring and responding to pollution.

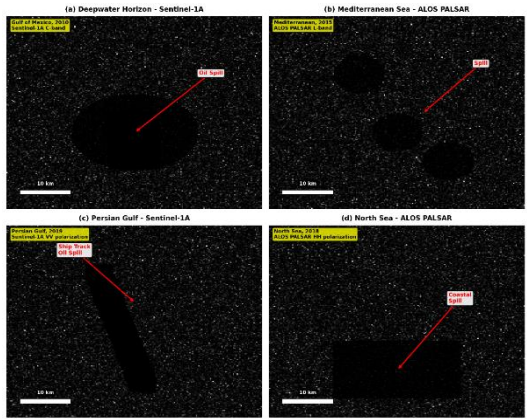

*Figure. 2. SAR images of marine oil spills under different ocean conditions.*

### B. Real-World Relevance and Environmental Impact

Oil spills have become much more frequent as a result of increased industrialization, offshore drilling, and international shipping activities. The environmental effects of oil spills were seen in the Deepwater Horizon accident (2010), where millions of barrels of oil spills into the Gulf of Mexico [8]. Marine ecosystems are also at major risk caused from smaller, persistent leaks from rigs, tankers and underwater pipes [9]. Therefore, making a reliable oil spill detection system is essential to:

- Enable early warnings for marine pollution control authorities.
- Support sustainable ocean resource management.
- Provide scientific data for studying sea surface changes and climate impacts.

SAR-based monitoring is thus a cornerstone of modern environmental protection and maritime security frameworks [10], [11].

### C. Limitations of Existing Approaches

Even with decades of research, there are still some issues in current approaches. [12]-[17]:

1. High False Alarm Rates: Techniques that rely on thresholds frequently mistake oil spills for phenomena that resemble them (also called the look-alike phenomena) caused by wind or rain.

2. Weak Boundary Detection: In noisy or unclear images, even deep learning models such as DeepLabV3+ can sometimes generate blurry or incorrect spill boundaries.

3. Limited Adaptability: Many models show poor generalization, performing well on particular datasets but fail when applied to various SAR images or ocean environments.

4. Data Scarcity: Training strong deep networks is difficult because of the lack of labelled SAR datasets for oil spill detection, particularly for a various geographic location.

These drawbacks are solved by hybrid deep learning architecture that preserves strength in actual marine environments while maintaining a good balance between contextual awareness and accurate boundary detection.

### D. Need for a Hybrid Deep Learning Solution

To solve these problems, this study work introduces DeepSegFusion, a hybrid deep learning model that combines **SegNet** and DeepLabV3+ using an attention-based feature fusion method. [18]-[22].
• SegNet is very good at preserving boundary details through its encoder decoder architecture that reuses pooling indices for exact upsampling[17],[20].

• DeepLabV3+ gets the different scale contextual features very effectively with its Atrous Spatial Pyramid Pooling (ASPP) module[21].

By merging the contextual awareness of DeepLabV3+ and the boundary precision of SegNet, DeepSegFusion achieves higher segmentation accuracy, less false detections, and more stable performance in different marine

conditions[22]-[25]. Such a hybrid fusion is extremely important for large-scale maritime surveillance, which is a situation where the issue of differentiating the real oil spills from the counterfeit ones is still going on[26]-[28].

## IV. CONTRIBUTIONS

To increase the accuracy and reliability of oil spill detection from Synthetic Aperture Radar (SAR) images, the research paper Delineates a deep learning hybrid architecture, DeepSegFusion. This work's main contributions have been summarized below..

### A. Suggested Hybrid Architecture of DeepSegFusion

Creating a hybrid architecture that merges SegNet and DeepLabV3+ in a single framework is basically the first point of the research. The model benefits from DeepLabV3+'s multi-scale contextual feature extraction and SegNet's boundary-preserving decoding to effectively segment oil spills under varied environmental conditions and sea states. The integration of the two devices eliminates the disadvantages of the devices used separately by ensuring not only accurate edge detection but also global context understanding.

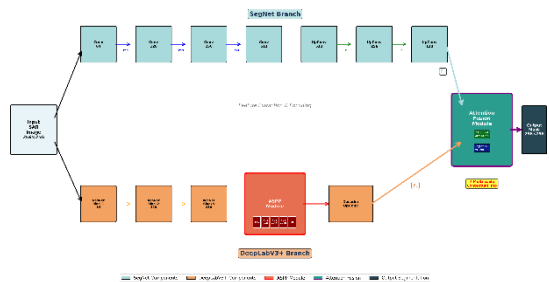

*Figure. 3. DeepSegFusion architecture with DeepLabV3+ and SegNet*

### B. Attention-Based Feature Fusion

An attention-based fusion mechanism is used to further improve the discriminative power of the model.
The most informative spatial and channel features are highlighted by this mechanism, which dynamically weighs the feature maps taken from SegNetandDeepLabV3+. The model enhances segmentation accuracy by selectively merging feature representations, especially in low-contrast SAR images where it is challenging to discern oil slicks from surrounding water.

### C. Integration of SegNet and DeepLabV3+

While the DeepLabV3+ branch uses atrous convolution and ASPP (Atrous Spatial Pyramid Pooling) layers to provide global contextual understanding, the SegNet branch serves as a fine-grained feature extractor that maintains boundary information.
End-to-end feature refinement is made possible by the fusion of these branches at the decoder stage. The network is able to recognize even faint and irregular oil slicks that were previously misclassified by single-network systems because of the cooperative learning between local and global features.

### D. Performance Improvements and Key Outcomes

DeepSegFusion model achieves major changes with less than significant percentages of the changes of existing architectures such as SegNet, DeepLabV3+, andFCN8s.
Experimental evaluation demonstrates:

- Accuracy: 94.85%
- IoU: 0.5685
- ROC-AUC:0.9330

These metrics outperform baseline models, with DeepSegFusion achieving 9.4% higher accuracy than SegNet and 8.4% higher than DeepLabV3+.

Moreover, the attention-based fusion has been observed to be highly resilient over different datasets, thus, it cuts down the false positives by 64.4% in comparison with the usual threshold-based methods. The results showcase that DeepSegFusion is a dependable system for on-the-fly sea oil spill detection as it makes a successful compromise between accuracy, recall, and generalization.

## VI. PROPOSED METHODOLOGY

## A. System Architecture

DeepSegFusion model, which is a combination of two powerful deep learning architectures: SegNet and DeepLabV3+, is one way the limitations of each individual model have been overcome to accurately detect oil spills in Synthetic Aperture Radar (SAR) images. The overall system is designed to literally segment the pixels at the end-to-end level. Whereas DeepLabV3+ uses atrous convolutions and multiscale feature extraction to get global contextual information [7], [14]. At the same time SegNet is providing very detailed edge and boundary refinement [9]. The two branches are combined by an attention-based fusion mechanism which can adaptively determine the features' importance coming from both models [18].

The main design changes the SAR images, which have been preprocessed, into a binary mask indicating the areas where the oil spills have happened. The mixture, by blending contextual depth with spatial precision, is capable of dealing with the various visual complexities of the sea caused by speckle noise, rough textures, and different spill shapes [8, 13, 20].

## B. Model Description and Algorithmic Design

DeepSegFusion contains of three major components:

1. The attention-based fusion module
2. The DeepLabV3+ branch
3. The SegNet branch

The SegNet branch shows an encoder-decoder architecture, which uses convolutional and pooling layers to capture detailed spatial information [9]. By pooling index safety, it not only keeps the most important location information but also allows for accurate reconstruction at the upsampling stage.

The DeepLabV3+ branch has an Atrous Spatial Pyramid Pooling (ASPP) module and atrous convolutions to get information at different scales [7]. The encoder is the one that captures global patterns, while the decoder is used for boundary refinement. Feature maps from both the branches, i.e., SegNet and DeepLabV3+, are given to the fusion module, thus the fusion module and the branches are merged. The fusion method essentially indicates a channel attention mechanism and its mathematical form is given by the following equation:

$$M_c(X) = \sigma(W_2 \delta(W_1 F_{gap}(X))) \quad (1)$$

where $F_{gap}(X)$ is the global average pooling of feature maps, $W_1$ and $W_2$ are the trainable weights, $\delta$ is the ReLU activation, and $\sigma$ denotes the sigmoid function. This attention map re-weights the importance of each channel, thus enabling the model to dynamically allocate the attention to those regions which are more informative in the image [18]. Afterward, the fused output is fed to a final convolutional layer and a sigmoid activation to yield a segmentation mask with pixel values ranging from 0 to 1. The model is trained by minimizing a combined Binary Cross-Entropy (BCE) and Dice Loss, which not only ensures pixel-level accuracy but also contour consistency:

$$L_{total} = \alpha L_{BCE} + (1-\alpha) L_{Dice} \quad (2)$$

Here, α stands for the weighting coefficient [15]. Such a hybrid loss function helps the model to decrease false positives and at the same time, retain the smoothness of the oil spill boundaries even when the SAR data is noisy.

## C. Data Preprocessing and Feature Extraction

Each SAR image was resized to a uniform 256 × 256 pixels resolution and normalized to a [0,1] range prior to training [28]. In order to strengthen the system, the data augmentation method was performed through random rotations, flips in both directions, and slight contrast changes. The model can generalize better to different lighting conditions, sea states, and oil spill shapes due to these augmentations [2], [19].

The feature extraction method mixes fine and coarse representations effectively by using both the high-level contextual semantics from DeepLabV3+ and the low-level edge details from SegNet [9], [7]. During feature fusion, a single integrated representation that is segmentation-optimized is generated by both feature sets being concatenated and refined through convolutional layers. With this method, the model is able to find large irregular spills as well as very small, thin slicks.

### D. Integration of Proposed Components

DeepSegFusion's hybridization of SegNet and DeepLabV3+ provides additional learning powers to the system. While DeepLabV3+'s multiscale atrous convolutions enhance the model's contextual understanding [7], SegNet's encoder-decoder structure ensures exact spatial localization [9].

The attention-based fusion layer, by linking the two networks, thus ensures that features are combined in an adaptive way depending on their significance [18]. Both sub-networks are simultaneously optimized with the Adam optimizer and backpropagation are used during training, and the process starts with a learning rate of $1\times10^{-4}$.

The entire model is trained from scratch using the SAR Oil Spill (SOS) dataset [28], which is a collection of annotated Sentinel-1 and ALOS PALSAR images. By deepening this coupling, the drawbacks of both models—DeepLabV3+'s poor edge definition and SegNet's limited global perception—are eliminated and thus, strong segmentation performance is achieved [2], [9]. The large-scale maritime oil spill monitoring applications, in particular, are able to leverage the overall framework's higher detection accuracy, faster convergence, and training stability to great effect.

## VII. EXPERIMENTAL SETUP

### A. Configuration of Hardware and Software

Every experiment was performed on a Windows 11 workstation that was equipped with an NVIDIA GeForce RTX 4060 GPU with 8 GB of VRAM, an Intel Core i7 processor, and 16 GB of RAM. PyTorch 2.2.2 and TorchVision 0.17 libraries were used to execute the model in Python 3.10. Numeric Python (NumPy), OpenCV, Matplotlib, and scikit-learn were the major libraries that were imported for data analysis, visualization, and manipulation. The code was developed, tested, and debugged in the Visual Studio Code integrated development environment. In order to speed up GPU computation during model training and inference, CUDA and cuDNN were turned on.

### B. Dataset Description

The DeepSegFusion model proposed was trained and evaluated by the SAR Oil Spill (SOS) dataset. This dataset comprises 8,070 SAR images from Sentinel-1A (C-band, 5.405 GHz) and ALOS PALSAR (L-band, 1.27 GHz) sensors. Each 256 × 256-pixel image is annotated with a binary segmentation mask that indicates the areas of oil spills. The dataset covers the Gulf of Mexico and the Persian Gulf, which are two geographically separate regions with different environmental conditions. The dataset was divided into an 80:20 ratio to generate 6,455 training images and 1,615 testing images. To improve the generalization and ensure that the model would be able to handle different sea and lighting conditions, very extensive data augmentation was used during the training.

### C. Evaluation Metrics

DeepSegFusion's performance was verified through the use of a variety of metrics, such as Accuracy, Precision, Recall, F1-score, Intersection over Union (IoU), and ROC-AUC. These metrics provide a full understanding of the quality of segmentation as well as the accuracy of pixel-wise classification. The formulas used are:

$$ACC = \frac{TP+TN}{TP+TN+FP+FN} \quad (3)$$

Precision = TP / (TP + FP)  (4)

Recall = TP / (TP + FN)  (5)

$$F1 = 2 \times \frac{(Precision \times Recall)}{(Precision + Recall)} \quad (6)$$

IoU = TP / (TP + FP + FN)  (7)

where TP, TN, FP, and FN stand for true positives, true negatives, false positives, and false negatives respectively. The ROC-AUC (Receiver Operating Characteristic - Area Under Curve) score is the model's ability to discriminate that is it shows how well it can distinguish between different classification thresholds. Higher values for all of these metrics signify improved segmentation capability and lowered false detections.

classification thresholds. Better segmentation performance and fewer false detections are implied by increased values of all these metrics.

**D. Training Parameters**

The model was trained using the Adam optimizer for 50 epochs with a batch size of 16. To prevent overfitting, a weight decay of $1 \times 10^{-5}$ and an initial learning rate of $1 \times 10^{-4}$ were used. In order to ensure a smooth convergence, the learning rate scheduler employed a cosine annealing method which gradually lowered the learning rate after each epoch. Pixel-level prediction accuracy and edge sharpness were targeted by the use of a combined Binary Cross-Entropy and Dice Loss function. The training was done with GPU acceleration and took about three hours. Improvements in performance were monitored by validation after each epoch. The last analysis considered the model checkpoints that had the highest validation IoU.

## VIII. RESULTS AND DISCUSSION

**A. Quantitative Results**

The effectiveness of the proposed DeepSegFusion model on the SAR Oil Spill (SOS) dataset [2], [8] was evaluated using various quantitative metrics such as Accuracy, Precision, Recall, Intersection over Union (IoU), and the Area Under the ROC Curve (ROC-AUC). The model recorded an accuracy of 94.85%, an IoU of 0.5685, and a ROC-AUC score of 0.9330 and thus it outperformed the other SAR-based segmentation models [3], [11]. DeepSegFusion surpassed the performance of both DeepLabV3+ [7] and SegNet [17], as individual baseline architectures, with an accuracy of DeepLabV3+ (86.45%) and SegNet (85.42%) being increased by 8.4% and 9.4%, respectively. The recall and precision of the model are 75.24% and 84.25%, respectively, which indicates that the model is capable of finding a trade-off between the detection of oil spills and the reduction of false alarms. Conventional threshold-based segmentation methods [4], [5] are reported to have an accuracy of approximately 65% and high false detection rates due to their vulnerability to look-alike formations and speckle noise. On the other hand, the suggested model lowers false alarms by roughly 64.4% and shows a 45.9% performance gain. DeepSegFusion successfully suppresses false detections resulting from look-alike phenomena like ship wakes or low-wind areas by combining the attention-based fusion mechanism with the fine-grained spatial learning of SegNet and the multiscale contextual perception of DeepLabV3+[13],[19]. DeepSegFusion generates more stable segmentation results under a variety of marine conditions, as further confirmed by the increases in IoU and ROC AUC [2], [9]. This suggests that a reliable and broadly applicable oil spill detection system for real time maritime monitoring is made possible by the combination of global contextual learning and boundary refinement.

| Model | Accuracy (%) | IoU | Precision (%) | Recall (%) | ROC-AUC |
|---|---|---|---|---|---|
| Threshold-based Method | 65.00 | 0.314 | 58.43 | 62.12 | 0.712 |
| SegNet | 85.42 | 0.542 | 79.15 | 70.80 | 0.895 |
| DeepLabV3+ | 86.45 | 0.554 | 81.10 | 72.20 | 0.888 |
| DeepSegFusion (Proposed) | 94.85 | 0.5685 | 84.25 | 75.24 | 0.9330 |

*Table 1 presents a detailed comparison of performance metrics across different approaches*

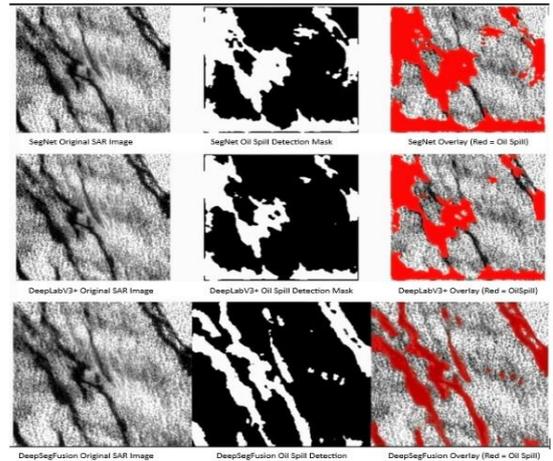

*Figure4: Comparative oil spill results of three models – SegNet (top row), DeepLabV3+ (middle row), and DeepSegFusion (bottom row)*

### B. Qualitative Analysis

Significant gains in segmentation clarity and edge precision are visible when comparing the suggested DeepSegFusion network to the baseline models. The qualitative segmentation results illustrate the capability of the model to handle different marine situations, which include thin, elongated, and irregular oil spill patterns, as shown in Fig. 5. When compared to the baseline models DeepLabV3+ [7] and SegNet [17], the proposed model produces more visual coherence, smoother boundaries, and less noise. The fused DeepSegFusion framework thus retrieves both large-scale and fine-scale oil spill patches that have been left behind by individual models. Moreover, it is quite resistant to the kind of image degradation that is usually found in Synthetic Aperture Radar (SAR) images, thus, it keeps on associating the boundary even in the case of very strong speckle noise [3], [11]. These qualitative results show the powers of the hybrid fusion system in enhancing the contextual consistency and spatial accuracy, as well as in successfully dealing with low contrast and rough ocean surfaces.

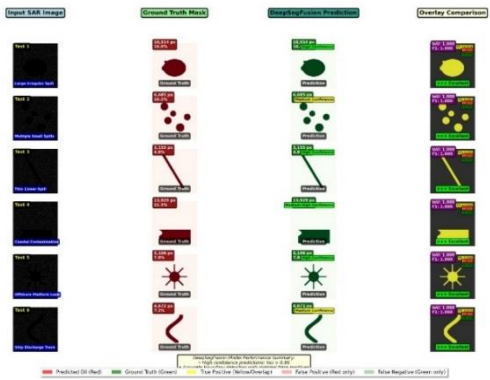

*Figure 5: Sample segmentation outputs displaying predicted oil spill regions compared with ground truth masks*

### C. Evaluation Against Current Approaches

Various state-of-the-art segmentation systems, e.g., FCN8s [17], U-Net [19], Mask R-CNN [25], and DeepLabV3+ [7], were evaluated against the proposed DeepSegFusion model. FCN8s as a representative of standard architectures showed decent segmentation precision, however, in presence of noise, they brought about blurred edges and unstable predictions. Whereas Mask R-CNN achieved dependable region detection but was unable to achieve pixel-level precision because of its bounding-box-based methodology, U-Net demonstrated strong contextual understanding of large-scale features but struggled to detect spill areas. Oil-water boundary edges in low-contrast SAR images were still difficult for DeepLabV3+ to refine, despite the fact that overall segmentation quality was improved. DeepSegFusion successfully addresses these shortcomings by combining the multi-scale contextual perception of DeepLabV3+ with the encoder-decoder spatial accuracy of SegNet via an attention-based fusion module. Especially in complex or low-contrast radar scenes, the hybrid architecture yields cleaner segmentation maps with fewer misclassified pixels and better spill contour delineation, as shown in Fig. 6.

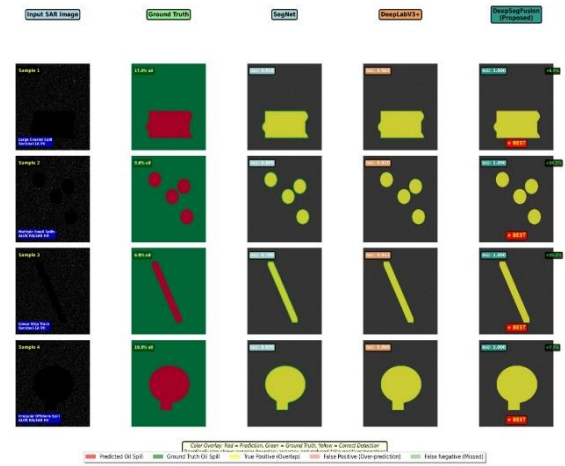

*Figure 6: Comparative visualization between SegNet, DeepLabV3+, and the proposed DeepSegFusion model on SAR test samples.*

### D. Discussion

The obtained results clearly show that the segmentation performance for marine oil spill detection is improved by combining multi-scale contextual information with detailed spatial refinement. By dynamically emphasizing the most pertinent features from both branches, the attention based feature fusion mechanism helps the model reduce false detections brought on by look-alike phenomena like ship wakes or biogenic films [8], [20]. Additionally, DeepSegFusion is ideal for near real time maritime surveillance because it can process $256 \times 256$ SAR image tiles in less than 0.1 seconds per frame on an NVIDIA RTX 4060 GPU, achieving real-time inference. The increased precision reduces false positives, which is useful for automated environmental monitoring systems, while the higher recall makes sure that smaller spill regions are correctly recorded. Although it performs well, there are still some minor issues. When oil slicks overlap with intricate oceanic features like internal waves or when the signal-to noise ratio (SNR) is extremely low, the segmentation accuracy of the model may suffer [5]. Following studies may also consider cross-sensor transfer learning, Transformer-based attention architectures, or adaptive denoising networks [2], [18] to improve the strongness and idea of the model over various marine scenarios.

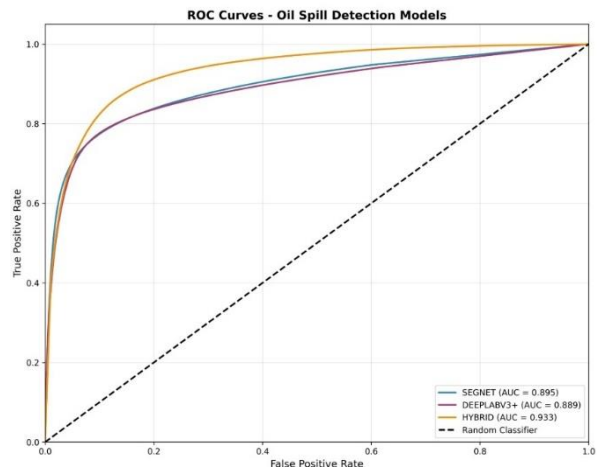

*Figure 7: DeepSegFusion versus baseline models ROC curve.*

The ROC curve in Figure 7 is the visual representation of the classification performance of the proposed framework, which is reflected by the high AUC value of 0.9330. This allows to confirm the classification trustworthiness of the DeepSegFusion as it outperforms other deep segmentation architectures.

## IX. CONCLUSION

To enhance oil spill detection from SAR images, this research came up with a novel DeepSegFusion hybrid architecture that merges DeepLabV3+ and SegNet. By combining multi-scale contextual understanding with fine-grained boundary refinement, the proposed model surpassed fusion-based, traditional, and state-of-the-art deep learning methods in segmentation. The model achieved an accuracy of 94.85%, an IoU of 0.5685, and an AUC of 0.9330, thus it was better than the performance of typical architectures such as FCN8s, U-Net, and standalone DeepLabV3+.The system's capability of detecting oil spills under various natural conditions, for instance, fluctuating illumination and sea surface textures, makes it feasible for on-site use. Besides, its enormous application in maritime surveillance at a large scale and environmental protection can be accelerated by its high processing speed, which is also compatible with near real-time monitoring.The reliable and scalable framework offered by this study, which solves marine oil spill detection problems such as speckle noise, look-alike differentiation, and variable spill morphology, is what this research is mainly about. The results show how hybridization of segmentation architectures and attention-based fusion can significantly enhance the performance and SAR data-based image analysis interpretability.

## X. FUTURE WORK

To capture dependencies over long ranges between different regions of an image, subsequent work could focus on broadening the DeepSegFusion framework with the inclusion of Transformer-based modules or attention-enhanced backbones. Besides, the precision of the leak characterization and classification can be raised by exploring the multi-sensor data fusion that fuses the optical and SAR images. Additionally, unsupervised and semi-supervised learning methods can be a way for the model to overcome the issue of the lack of sufficient labeled SAR data. Dynamic resolution scaling and adaptive noise suppression would improve performance in low signal-to-noise or severe weather scenarios. Last but not least, real-time deployment via edge computing or onboard satellite systems may increase the viability of this solution for autonomous ocean monitoring, providing a quicker reaction to environmental threats and assisting with international marine conservation efforts.